# Enhancing Graph Representation Learning with Attention-Driven Spiking Neural Networks

**Huifeng Yin [1], Mingkun Xu [1], Jing Pei [1] and Lei Deng [1]**

[1] Center for Brain-Inspired Computing Research (CBICR), Department of Precision Instrument,
Tsinghua University, Beijing, China
E-mail: {peij,leideng}@mail.tsinghua.edu.cn

**Abstract:** Graph representation learning has become a crucial task in machine learning and data mining due to its potential for modeling complex structures such as social networks, chemical compounds, and biological systems. Spiking neural networks (SNNs) have recently emerged as a promising alternative to traditional neural networks for graph learning tasks, benefiting from their ability to efficiently encode and process temporal and spatial information. In this paper, we propose a novel approach that integrates attention mechanisms with SNNs to improve graph representation learning. Specifically, we introduce an attention mechanism for SNN that can selectively focus on important nodes and corresponding features in a graph during the learning process. We evaluate our proposed method on several benchmark datasets and show that it achieves comparable performance compared to existing graph learning techniques.

**Keywords:** Graph representation learning, Attention mechanisms, Spiking neural networks, Graph neural networks.

## 1. Introduction

Graphs are ubiquitous data structures that can represent a wide range of complex systems, from social networks to biological networks. Graph representation learning seeks to create effective embeddings that encapsulate a graph's structural and semantic information, facilitating tasks like node classification, edge classification, and graph classification. Recently, spiking neural networks (SNNs) have been proposed as a powerful approach to graph representation learning due to their ability to encode and process spatiotemporal information in a more biologically plausible and energy-efficient way than traditional neural networks [1].

Recently, various approaches have been proposed for graph representation learning, including spectral methods, random walk-based methods, and neural network-based methods [2]. Among these, neural network-based methods, such as graph convolutional networks (GCNs) [3] and graph attention networks (GATs), have gained significant attention due to their capacity to capture complex nonlinear relationships in graph structures. However, traditional neural networks have limitations in processing spatiotemporal information, which is essential for numerous graph learning tasks due to the human brain's constant interaction with dynamic stimuli and evolving environments. SNNs, inspired by the way neurons communicate in the brain through spikes, are capable of encoding and processing spatiotemporal information more efficiently and in a biologically plausible manner, making them an attractive alternative for graph representation learning. But they still face challenges in processing large-scale graphs efficiently.

To address this issue, attention mechanisms have been introduced to selectively focus on relevant parts of the node features, allowing the network to selectively attend to important nodes and corresponding features in the graph [4]. In this paper, we propose an attention-driven SNN model that combines the benefits of attention mechanisms with the efficiency and interpretability of SNNs for graph representation learning. We demonstrate the effectiveness of our proposed model through several experiments on various benchmark datasets, showing comparable performance and better biological plausibility compared to existing graph representation learning methods.

## 2. Methods

Our proposed Spiking Graph Attention Network (SpikingGAT) model combines attention mechanisms with Graph Spiking Neural Network (Graph-SNN) to enable efficient graph representation learning [1]. Specifically, we introduce a graph attention mechanism that computes attention coefficients for each pair of nodes, allowing the model to weigh the contributions of neighboring nodes appropriately. This mechanism enables the SNN to focus selectively on relevant nodes and their corresponding features and effectively capture the underlying graph structure, during the learning process, as shown in Fig.1. By integrating attention mechanisms into the Graph-SNN architecture, our SpikingGAT model achieves superior graph representation learning outcomes.

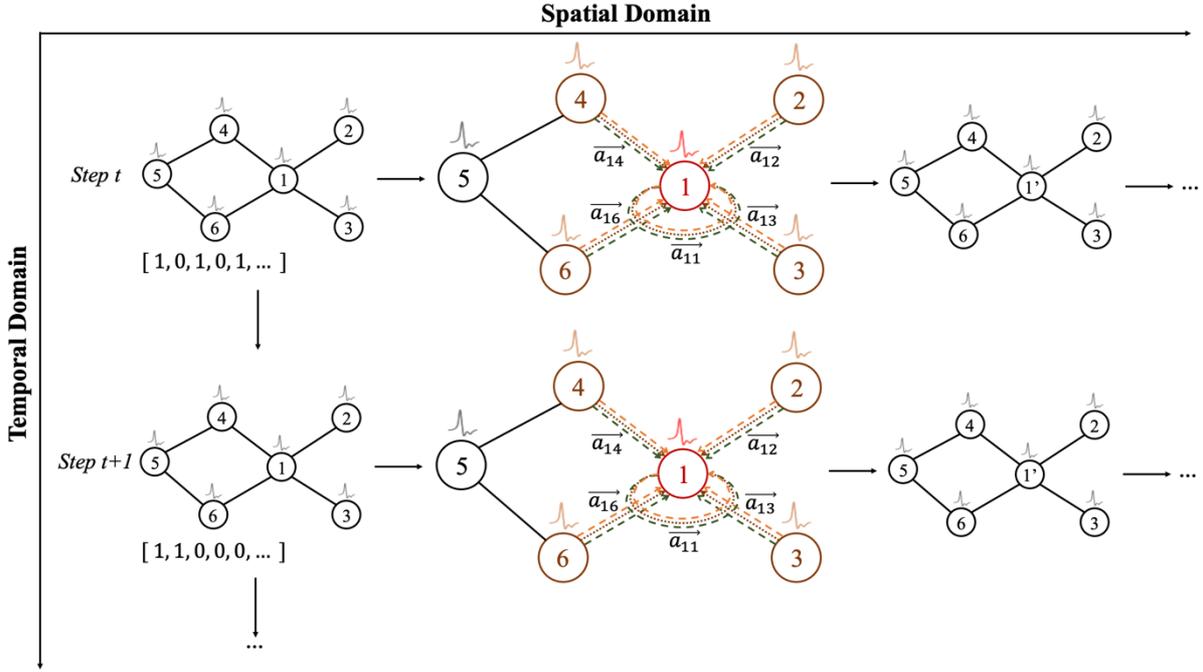

**Fig. 1.** The schematic diagram of SpikingGAT, showcasing the integration of multi-head attention and temporal domain within SNNs. Various arrow styles and colors signify distinct attention heads being performed independently with K = 3 heads.

Specifically, in our SpikingGAT model, we first compute the attention coefficients for each node pair by applying a compatibility function that takes into account the nodes' features and their relative positions in the graph. The attention coefficient between node $i$ and node $j$ for the $k^{th}$ head is computed as follows:

$$e_{ij}^{(k)} = \text{LeakyReLU}\left(\mathbf{a}^{(k)T}[\mathbf{W}^{(k)}\mathbf{h}_i \| \mathbf{W}^{(k)}\mathbf{h}_j]\right) \quad (1)$$

where $\mathbf{W}^{(k)}$ is a learnable weight matrix for the $k^{th}$ head, $\mathbf{h}_i$ and $\mathbf{h}_j$ are the feature vectors of nodes $i$ and $j$, respectively, $\mathbf{a}^{(k)}$ is a learnable attention vector, and LeakyReLU is the activation function. These coefficients are then normalized using a softmax function to ensure that they sum to one, promoting a smooth distribution of attention weights among neighboring nodes:

$$\alpha_{ij}^{(k)} = \frac{\exp\left(e_{ij}^{(k)}\right)}{\sum_{k \in \mathcal{N}_i} \exp\left(e_{ik}^{(k)}\right)} \quad (2)$$

where $\mathcal{N}_i$ is the set of neighboring nodes of node $i$, and $\alpha_{ij}^{(k)}$ is the normalized attention coefficient for the $k^{th}$ head.

Once we have computed the attention coefficients for each node pair, we apply the multi-head attention mechanism to the input features of each node by multiplying them with the corresponding coefficients. This operation is carried out at each layer of the SpikingGATs, updating the node representations with attention-weighted information from their neighbors. For each head $k$, the output feature matrix is given by:

$$\mathbf{H}^{(l+1,k)} = \sigma\left(\sum_{j=1}^{N} \alpha_j^{(k)} \mathbf{W}^{(l,k)} \mathbf{H}^{(l)}\right) \quad (3)$$

where $\sigma$ is a nonlinear activation function, such as the ReLU function. Then, the output feature matrices from all $k$ heads are concatenated to form the final output feature matrix:

$$\mathbf{H}^{(l+1)} = \text{concat}\left(\mathbf{H}^{(l+1,1)}, \mathbf{H}^{(l+1,2)}, \ldots, \mathbf{H}^{(l+1,K)}\right) \quad (4)$$

To account for the spiking behavior in SNNs, we employ a Leaky Integrate-and-Fire(LIF) [5] neuron model, which integrates the input features weighted by attention over time. The membrane potential is described by the following differential equation:

$$\tau \frac{du}{dt} = -[u(t) - u_{\text{rest}}] + RI(t) \quad (5)$$

where $\tau$ is the membrane time constant, $u_{rest}$ is the resting membrane potential, $R$ is the membrane resistance, and $I(t)$ represents the input current. When the membrane potential exceeds a certain threshold value $u^{th}$, the neuron generates a spike and the membrane potential is reset to the resting potential $u_{rest}$ This process can be formulated as:

$$\begin{cases} o^t = 1, u^t = u_{rest} & \text{if } u^t \geq u_{th} \\ o^t = 0 & \text{if } u^t < u_{th} \end{cases} \quad (6)$$

where $o^t$ represents the output spike at the time step $t$.

The SpikingGAT model architecture consists of multiple layers of LIF neurons, with each layer responsible for aggregating and transforming the attention-weighted features of the neighboring nodes.

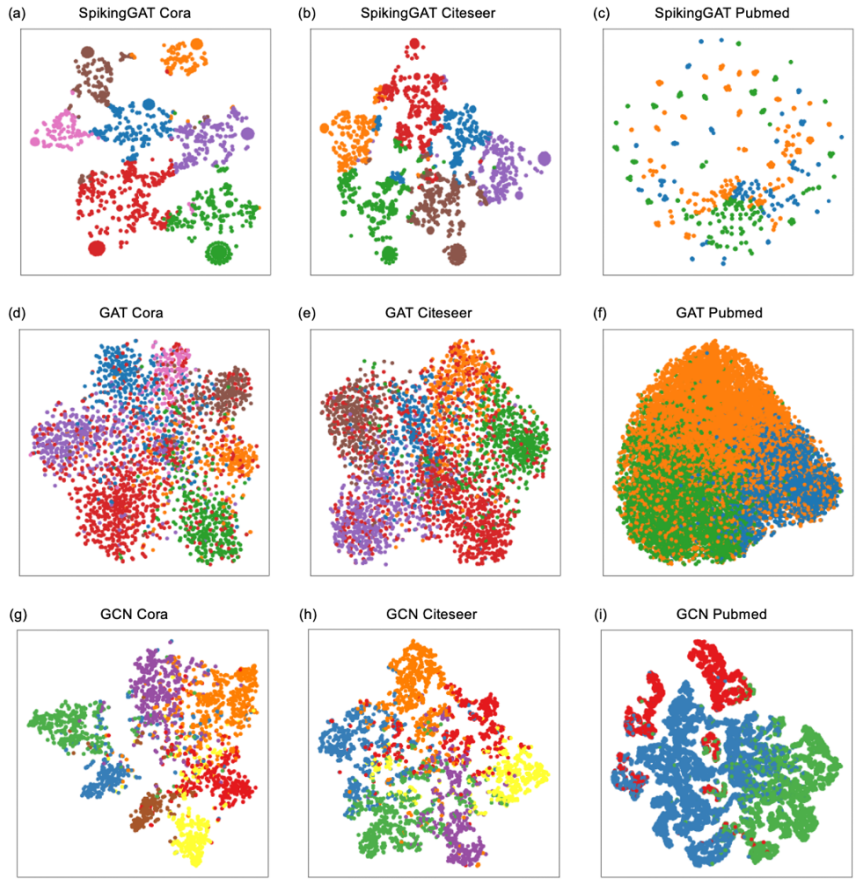

**Fig. 2.** (a)(b)(c)Visualization of the last hidden layer features of the SpikingGAT on the Cora, Citeseer, and Pubmed datasets produced by t-SNE. (d)(e)(f)Visualization of the last hidden layer features of the GAT on the Cora, Citeseer, and Pubmed datasets produced by t-SNE. (g)(h)(i) Visualization of the last hidden layer features of the GCN on the Cora, Citeseer, and Pubmed datasets produced by t-SNE.

The output of the final layer is a set of effective embeddings that capture both the graph structure and node features.

We trained our SpikingGAT model using the iterative spiking message passing [1, 6] method of SNN that adjusts the connection weights between neurons based on the spike propagation and attention mechanism. This learning rule enables the network to adapt its connection strengths in response to both the graph structure and node features. Additionally, incorporating attention mechanisms enhances the model's ability to learn graph representations, focusing on the most informative connections in the graph.

### 3. Experiments

#### 3.1. Basic experiments

To evaluate the performance of our proposed SpikingGAT model, we first conducted experiments on three standard citation datasets, Cora, Citeseer, and Pubmed, where nodes represent paper documents and edges are undirected citation links. We provide a summary of the dataset statistics in Table 1, which is utilized to assess the model's ability to learn meaningful node representations within an single graph. [3].

**Table 1.** Overview of citation datasets

| Dataset | Cora | Citeseer | Pubmed |
|---|---|---|---|
| Nodes | 2708 | 3327 | 19717 |
| Edges | 5429 | 4732 | 44338 |
| Node feat. | 1433 | 3703 | 500 |
| classes | 7 | 6 | 3 |
| Training/Validation/ Testing | 140/500 /1000 | 120/500 /1000 | 60/500/ 1000 |

The Cora dataset contains 2708 nodes, 5429 edges, 7 classes, and 1433 features per node. The Pubmed dataset contains 19717 nodes, 44338 edges, 3 classes, and 500 features per node. The Citeseer dataset contains 3327 nodes, 4732 edges, 6 classes, and 3703 features per node. Each document node has a class label. We only use 20 labels per class during training with all feature vectors.

In these experiments, we maintain consistent settings for models across each dataset to ensure fairness. We employ the Adam [7] optimizer, setting an initial learning rate of 0.01 for GCN, and 0.005 for GAT and our SpikingGAT. All models are executed for 200 epochs, and we conduct 10 trials with varying random seeds. In each trial, the models are initialized using a uniform initialization method and trained by minimizing the cross-entropy loss on the training nodes. From the perspective of SNN, we set the time window ($T$) to 8 and establish a threshold ($u^{th}$) of 0.25 for basic performance evaluation. Specially, we set the leakage factor of 1, which does not exhibit decay in historical membrane potential as the time step progresses. For GATs and SpikingGATs, we adopt an MLP structure [Input-64-Output] with 8 attention heads. For GCNs, we adopt an MLP structure [Input-400-16-Output]. We implemented these models using the PyTorch deep learning framework and the Deep Graph Library(DGL). We trained the model on GPU-enabled hardware for efficient parallel processing of the spiking neurons and attention mechanisms.

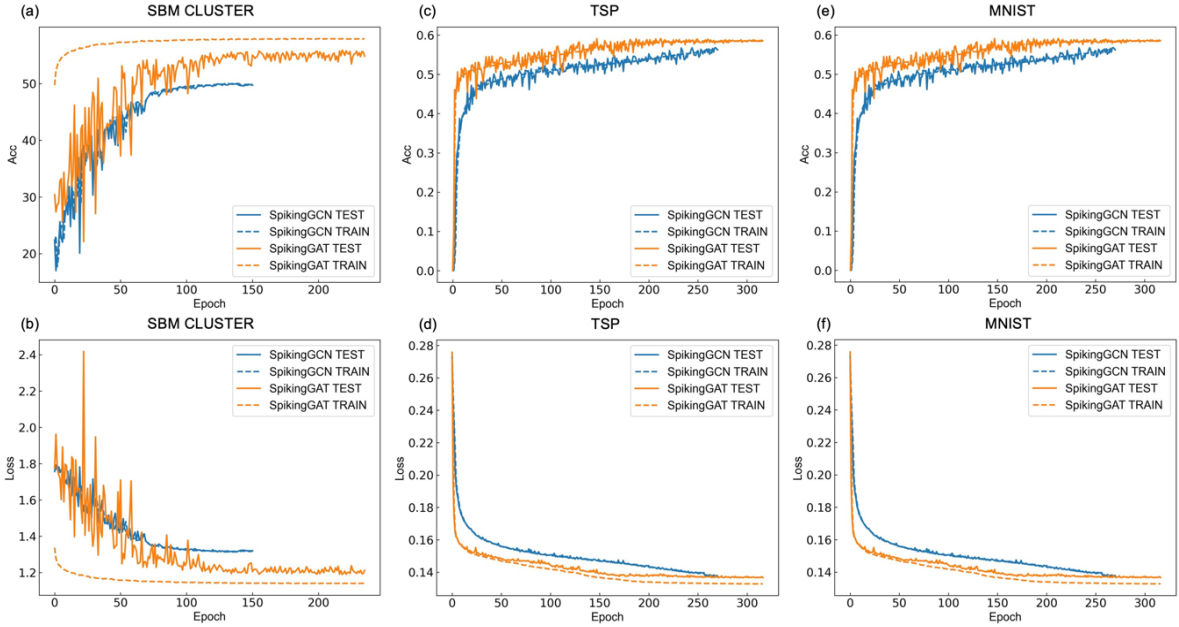

**Fig. 3.** (a)(c)(e) Accuracy variation of SpikingGAT and SpikingGCN with respect to the epoch on training and test dataset. (b)(d)(f) Loss variation of SpikingGAT and SpikingGCN with respect to the epoch on training and test dataset.

Our results from 10 trials are presented in Table 2. The results indicate that, despite employing binary spiking communication, our SpikingGAT models achieve performance comparable to the state-of-the-art results with a slight gap. This demonstrates the feasibility and capability of the spiking mechanism and spatiotemporal dynamics in handling diverse features from different nodes, as well as their effectiveness in graph scenarios with few labels.

**Table 2.** Performance comparison on citation datasets

| Method | Cora (ACC.±s.d.) | Citeseer (ACC.±s.d.) | Pubmed (ACC.±s.d.) |
|---|---|---|---|
| GCN [4] | 81.4±0.5 | 70.9±0.5 | 79.0±0.3 |
| GAT [4] | 83.0±0.7 | 72.5±0.7 | 79.0±0.3 |
| **SpikingGAT** | **79.9±0.9** | **68.4±0.5** | **78.0±0.5** |

Furthermore, we employed t-SNE [8] to visualize the feature representation capabilities of our SpikingGAT model compared to GCN and GAT in their last hidden layer features, as shown in Fig.2. Specifically, on the Cora, Citeseer, and Pubmed datasets, our SNN model demonstrated superior performance in distinguishing between different classes and convergence within the same class, thereby demonstrate more effective representation learning capabilities compared to GCN and GAT.

### 3.2. Extended experiments

In addition to the basic experiments on single graph datasets, we conducted extended experiments on multi-graph datasets to further validate the effectiveness and versatility of our SpikingGAT model in various graph learning tasks. By evaluating our model on tasks such as node classification, edge classification, and graph classification, we aimed to demonstrate the model's capability to generalize and adapt to different types of graph tasks. We summarize these dataset statistics used in extended experiments in Table 3.

**Table 3.** Overview of multi-graph datasets [9]

| Dataset | SBM CLUSTER | TSP | MNIST |
|---|---|---|---|
| Graphs | 12000 | 12000 | 70000 |
| Avg. Nodes | 117.20 | 275.76 | 70.57 |
| Avg. Edges | 4301.72 | 6894.04 | 564.53 |
| Node feat. | Node Attr(7) | Coord(2) | Pixel+Coord(3) |
| classes | 6 | 2 | 10 |
| Training/Validation/Testing | 10000/1000/1000 | 10000/1000/1000 | 55000/5000/10000 |
| Task Type | Node Classification | Edge Classification | Graph Classification |

Specifically, for node classification tasks, we used the SBM CLUSTER datasets, which are generated with the Stochastic Block Model(SBM). SBM is a traditional graph generation model in which each node belongs to a different community, and each community is connected with different probabilities [9].

For edge classification tasks, we employed the TSP dataset, which is about the traveler's problem, as follows: "Given a list of cities and the distances between each pair of cities, what is the shortest possible route that visits each city and returns to the origin city?". This dataset is a collection of complete graphs with weighted edges representing the distances between cities. The goal of this task is to classify edges as part of the optimal tour or not [10], testing our model's capacity to learn edge-level representation effectively.

Finally, for graph classification tasks, we convert each image in the popular MNIST datasets into graphs using super-pixels and classify these graphs. The node features of the graph are generated by the intensity and position of the super-pixels, and the edges are k nearest neighbor super-pixels, which is set to 8 [9]. The resulting graphs are of sizes 40-75 nodes. In this experiment, we aimed to evaluate the SpikingGAT model's performance in learning meaningful representations for entire graphs, capturing the global structure and relationships between nodes.

In these extended experiments, We employ the Adam optimizer, setting the initial learning rate of 0.001 and the minimum learning rate of $10^{-5}$. All models run up to 500 epochs, and we conduct 4 trials with varying random seeds. From the perspective of SNN, we maintained the same experimental settings as in the basic experiments with the time window $T = 8$, the firing threshold $V^{th} = 0.25$, and the leakage factor of 1. For GAT and SpikingGAT, we adopt 4 hidden layers of 19 dimension with 8 attention heads and an MLP layer (152 neurons) for classification. For GCN and SpikingGCN, 4 hidden layers of 152 dimension and an MLP layer of 152 neurons.

As shown in Table 4, our SpikingGAT models outperform the GCN models in SBM CLUSTER and MNIST datasets and outperform the SpikingGCN models in all datasets. And in Fig.3, we plotted the accuracy and loss curves of SpikingGAT and SpikingGCN during both the training and testing phases. The curves illustrate that SpikingGAT converges faster than SpikingGCN and achieves higher accuracy and lower loss during both the training and testing phases. The results demonstrate the effectiveness of incorporating attention mechanisms into Spiking Neural Networks to enhance their performance in different graph learning tasks, highlighting their generalization capabilities.

Besides, our SpikingGAT models can achieve comparable performance with the GAT models with a minor gap. It proves the feasibility and capability of spiking mechanism and spatial-temporal dynamics, which can work well on graph representation learning.

In this section, we have evaluated the performance of our proposed SpikingGAT model on a range of graph learning tasks, including node classification, edge classification, and graph classification. Our experiments on both single graph datasets and multi-graph datasets have demonstrated the SpikingGAT model's capability to learn effective representations and generalize to different types of graph tasks.

**Table 4.** Performance comparison on multi-graph datasets

| Method | SBM CLUSTER (ACC.±s.d.) | TSP (F1 SCORE ±s.d.) | MNIST (ACC.±s.d.) |
|---|---|---|---|
| GCN [9] | 47.828 ±1.510 | 0.643 ±0.001 | 90.120 ±0.145 |
| GAT [9] | 57.732 ±0.323 | 0.671 ±0.002 | 95.535 ±0.205 |
| Spiking GCN | 50.181 ±1.284 | 0.568 ±0.005 | 92.318 ±0.005 |
| **Spiking GAT** | **55.576 ±0.193** | **0.589 ±0.005** | **95.483 ±0.002** |

## 4. Conclusion

This paper introduces a novel Spiking Graph Attention Network (SpikingGAT) model that effectively integrates attention mechanisms with spiking neural networks for graph representation learning. SpikingGAT can effectively deal with the spatiotemporal information in graph structures, which is a limitation of traditional neural networks. In addition, the integration of attention mechanisms allows the model to selectively focus on important nodes and features, resulting in improved performance in graph representation learning tasks.

We evaluate the performance of our SpikingGAT model on various benchmark datasets, including single graph datasets (Cora, Citeseer, and Pubmed) and multi-graph datasets (SBM CLUSTER, TSP, and MNIST). Our experiments demonstrate that SpikingGAT model achieves comparable performance to GCN and GAT models while maintaining better biological plausibility. Furthermore, our model exhibits better performance in graph representation learning, as evidenced by t-SNE visualizations.

These experiments validate the effectiveness and versatility of our SpikingGAT model, demonstrating its ability to generalize and adapt to different types of graph tasks. By combining efficient attention mechanisms with interpretable SNNs, our work opens up new possibilities for future research and applications in graph representation learning.

## Acknowledgements

This work was supported by Science and Technology Innovation 2030 - New Generation of Artificial Intelligence, China project (2020AAA0109101), National Natural Science Foundation of China (No. 62106119, 62276151) and Zhejiang Lab's International Talent Fund for Young Professionals.